\documentclass[letterpaper, 10 pt, conference]{ieeeconf}

\IEEEoverridecommandlockouts                              

\usepackage{xcolor}

\title{\LARGE \bf
OTTR: Off-Road Trajectory Tracking using Reinforcement Learning
}

\author{Akhil Nagariya$^{1}$, Dileep Kalathil$^{2}$, Srikanth Saripalli$^{3}$
\thanks{$^{1}$Akhil Nagariya and $^{3}$Srikanth Saripalli are with the Department of Mechanical Engineering, Texas A\&M University,
        College Station, Texas 77845, USA
        {\tt\small akhil.nagariya@gmail.com, ssaripalli@tamu.edu}}
\thanks{$^{2}$Dileep Kalathil is with the Faculty of Electrical \& Computer Engineering, Department of Electrical \& Computer Engineering, Texas A\&M University, College Station, Texas 77845, USA
        {\tt\small dileep.kalathil@tamu.edu}}%
}
\usepackage{amsmath}
\usepackage{algorithmicx}
\usepackage{algorithm}
\usepackage{algpseudocode}
\usepackage{amssymb}
\usepackage{graphicx}
\usepackage[font=scriptsize]{caption}
\usepackage{subcaption}
\usepackage{bm}
\usepackage{float}
\usepackage{hyperref}

\usepackage{etoolbox}
\makeatletter
\AtBeginEnvironment{thebibliography}{%
   \clubpenalty10000
   \@clubpenalty \clubpenalty
   \widowpenalty10000}
\makeatother

\begin{document}

\maketitle	
\thispagestyle{empty}
\pagestyle{empty}

\begin{abstract}


In this work, we present a novel Reinforcement Learning (RL) algorithm for the off-road  trajectory tracking problem. Off-road environments involve varying terrain types and elevations, and it is  difficult to model the interaction dynamics of specific off-road vehicles with such a diverse and complex environment. Standard RL policies trained on a simulator will fail to operate in such challenging real-world settings. Instead of using a naive domain randomization approach, we propose an innovative supervised-learning based approach for overcoming the sim-to-real gap problem. Our approach efficiently exploits the limited real-world data available to adapt the baseline RL policy obtained using a simple kinematics simulator.  This avoids the need for modeling the diverse and complex interaction of the vehicle with off-road environments. We  evaluate the performance of the proposed algorithm  using two different off-road vehicles, Warthog (Fig. \ref{fig:warthog}) and Moose (Fig. \ref{fig:moose}). Compared to the standard ILQR approach,  our proposed approach  achieves a 30\% and 50\% reduction in cross track error in Warthog and Moose, respectively, by utilizing only 30 minutes of real-world driving data.


\end{abstract}


\section{Introduction} 
\label{intro}
\begin{figure}[H]
   \centering{
   \vspace*{-3cm}
   \resizebox{85mm}{87mm}{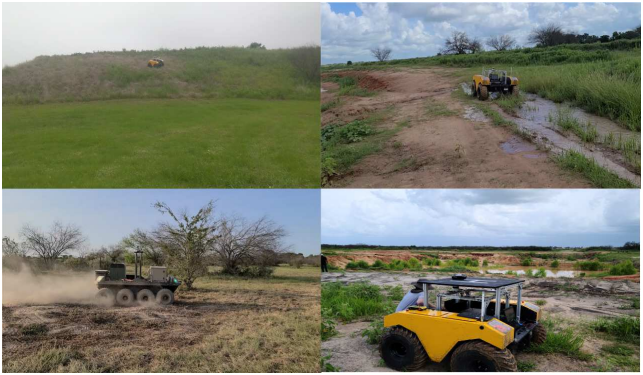}
   \caption{Shows various terrains involved in off-road testing of Moose and Warthog. The 
   test scenarios include grass, mud, gravel, dust along with elevation changes.}
   \label{fig:terrains}
   }
 \end{figure}

Trajectory tracking is a classical problem in robotics control, where a robot  is required to follow a trajectory specified by a sequence of waypoints  while satisfying various dynamic constraints on its  acceleration, linear velocity, angular velocity, torque limits etc. Trajectory tracking problem is particularly difficult  in the off-road environments, which involve various non-uniform terrain types and elevation changes as shown in Fig. \ref{fig:terrains}. Moreover, it is extremely challenging to characterize the interaction dynamics of specific off-road vehicles (often, with proprietary model) with such a diverse and complex environment. So, the classical model-based approaches for trajectory tracking are less effective for the off-road settings. A learning-based approach, which can avoid the direct modeling of the vehicle-environment interaction dynamics  using  data samples, is an attractive candidate for solving the off-road trajectory tracking problem. 

Reinforcement Learning (RL) is a class of machine learning that focuses on learning the optimal control policy when the precise model of the dynamical  system is unknown. Model-free RL algorithms have seen breakthrough successes recently in a number of application such as playing games \cite{mnih2015human, silver2016mastering} and robotics control  \cite{lillicrap2016continuous, levine2016end, akkaya2019solving, haarnoja2019learning}.  However, most of these successes are either in  simulation domain or in structured real-world settings, which are  significantly different from the challenging off-road real-world environments. Model-free RL algorithms also require a very large number of data samples to converge to a reasonable policy. Training RL algorithms in the real-world setting is infeasible because it can be catastrophic; for example, in high speed off-road vehicle. The data requirement is typically achieved by using a simulator  to generate samples. However, it is very difficult to incorporate the diverse and complex off-road environment models into a standard simulator. This leads to the problem known as simulation-to-reality (sim-to-real) gap, where the RL policies learned using a standard simulator may not work well for the real-world setting.

\begin{figure}[H]
      \centering{
  \vspace*{-3cm}
  \resizebox{80mm}{75mm}{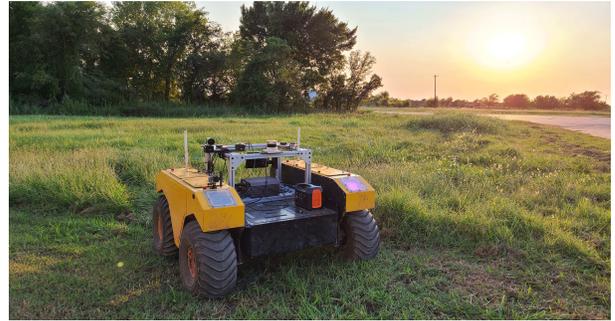}
  \caption{Warthog is an off-road differential drive vehicle with dimension 1.52 x 1.38 x 0.83 m, four wheels, mass 280kg and max payload capacity of 272kg \cite{warthog}.}

  \label{fig:warthog}
}
\end{figure}
\begin{figure}[H]
  \centering{
  \vspace*{-4.8cm}
  \resizebox{80mm}{90mm}{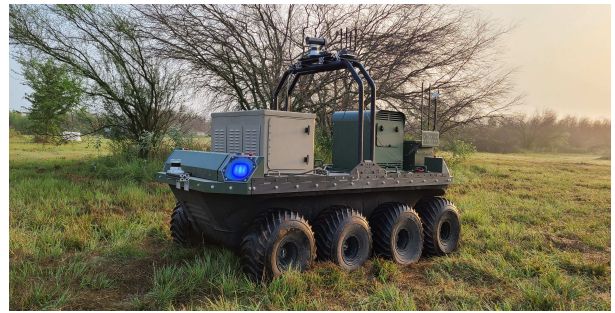}
 \caption{Moose is an off-road differential drive vehicle with dimension 3 x 1.5 x 1.1 m, eight wheels, mass 1590kg and max payload capacity of 513kg \cite{moose}. }
 \label{fig:moose}
}

\end{figure}

 Domain randomization \cite{sadeghi2017cad, tobin2017domain, peng2018sim} is the standard approach used for overcoming the sim-to-real gap,  which involves randomizing the dynamics and the simulation parameters during the learning. Domain randomization, however,  requires a sophisticated simulation model for the randomization process, and the unavailability of the proprietary information (of vehicles and simulators)  makes this approach less effective for  the off-road trajectory tracking problem. Moreover, domain randomization produces overly conservative policies which might also affect the performance.

 In this work, we propose a novel algorithm to overcome the sim-to-real gap for the off-road trajectory tracking problem by exploiting the limited  real-world data available. Our proposed algorithm uses the limited available data to modify the baseline policy obtained using an RL algorithm trained on a simulator. Our approach only requires  a simple kinematics simulator that avoids the need for modeling the  complex interaction of the vehicle with off-road environments. We collect the data and  evaluate the performance of the proposed algorithm  using two different off-road vehicles, Warthog (Fig. \ref{fig:warthog}) and Moose (Fig. \ref{fig:moose}). Compared to the standard iterative linear quadratic regulator (ILQR) approach, our algorithm achieves a 30\% and 50\% reduction in crosstrack error in Warthog and Moose, respectively, by utilizing only 30 minutes of real-world driving data.

\section{related work}
\label{rwork}
%

Model-based  approaches rely on learning the dynamic model of the system to develop the controller.  \cite{80202} and \cite{nolin2} are some of the earliest works that used a neural network for modeling the dynamics of the system and developing a controller based on the learned model. \cite{IEEE:1383790,IEEEexample:gprl2, IEEEexample:localgp, IEEEexample:gpquad, IEEEexample:gpdataeff, IEEEexample:gprldataeff} use a Gaussian process to model low dimensional stochastic dynamical system where few data points are available. They addressed model learning for robots in simulation \cite{ IEEE:1383790, IEEEexample:gprldataeff}, UAVs \cite{IEEEexample:gprl2, IEEEexample:gpquad}, and manipulators  \cite{IEEEexample:gpdataeff}. In our prior work  \cite{DBLP:abs200714492}, we proposed an approach that used neural networks to learn the dynamic model of the off-road and on-road vehicles and validate the learned model
by integrating it with a controller for trajectory tracking.

Model-free algorithms using simulators are also popular in robotics, \cite{PETERS2008682}, \cite{DBLP:KoberP08}, \cite{5649089},
\cite{DBLP:KoberOP11}, \cite{DBLP:KoberOP11}, especially for tasks that are not safety-critical. Since learning online in the real-world setting is expensive and often infeasible due to safety concerns, domain randomization is used as popular approach for overcoming the sim-to-real gap. \cite{tobin2017domain}  used domain randomization in robotics for object localization in the scene. \cite{zakharov2019deceptionnet} used a task network as the adversarial guide to modify 
the environment instead of direct domain randomization while \cite{tobin2018domain} proposed a data generation pipeline
capable of generating different kinds of object in simulation and training a deep neural network 
to perform grasp planning on these objects.
\cite{peng2018sim} used  domain randomization for overcoming the sim-to-real gap due to imperfect dynamic models. \cite{chebotar2019closing} proposed to learn a good randomization strategy for the dynamics parameters by online data collection during learning. These approaches have been shown to work on robotic manipulators and toy cars where a detailed simulator model is available from  the manufacturer. However, as mentioned before, the unavailability of the proprietary information (of vehicles and simulators)  makes this approach less effective for the off-road trajectory tracking problem. 


\begin{figure}[H]
  \centering{
  \vspace*{-3.0cm}
  \resizebox{70mm}{!}{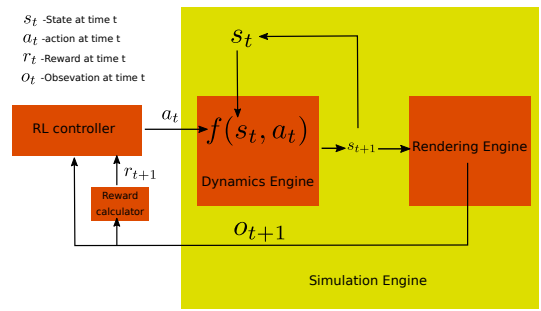}
  \caption{Simulation Engine}
  \label{fig:sim_eng}
  }
\end{figure}


\section{baseline rl policy for trajectory tracking}

   We define the trajectory by a set of $n$ ordered waypoints, $\{w_i = (x_{i}, y_{i}, \theta_{i}, v_{i})\}_{i=1}^{n}$, where  $(x_{i},y_{i})$ is the Euclidean coordinates, $\theta_{i}$ is the  desired vehicle orientation, and $v_{i}$ is the  desired vehicle velocity of waypoint $i$.  Given a trajectory, our goal is to learn a trajectory tracking controller that can minimize the crosstrack error in real-world off-road environments with two different offroad vehicles, Warthog and  Moose. While  both of these vehicles follow differential drive kinematics and accept linear and angular velocities as control commands, their dimension, mass and load distribution are quite different as specified in Fig. \ref{fig:warthog} and Fig. \ref{fig:moose}. Irrespective of these differences in the models, we train a single baseline RL policy for trajectory tracking for these vehicles. This RL policy is trained on a simulator that only simulates the kinematics and ignores complex tire-terrain interactions involved in the off-road scenarios and dynamics of these vehicles. In the next section, we will describe our approach to adapt this baseline RL policy for specific vehicles using limited real-world data from the corresponding  vehicles.

We use the simulation engine shown in Fig. \ref{fig:sim_eng} for training the baseline RL policy. It consists of the following two components:  the dynamics engine which simulates the forward kinematics of a differential drive vehicle, and the rendering 
engine which generates the observation given the current state of the vehicle. We use the standard gym interface from OpenAI Gym \cite{DBLP:journals/corr/BrockmanCPSSTZ16} to develop this simulation engine.
   
   

 \begin{figure}[H]
  \centering{
  \vspace*{-2.9cm}
  \hspace*{-1.5cm}
  \resizebox{70mm}{!}{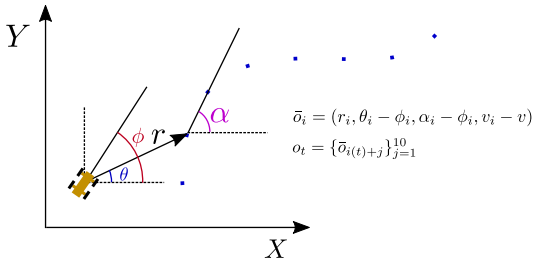}
  \caption{The blue dots represent waypoints. $o_i$ is the observation corresponding to the $i$th waypoint. The complete observation is represented by $o_t$ which consists of the observations
  corresponding to the closest 10 waypoints.}
  \label{fig:warobs}
  }
  
\end{figure}

The RL policy takes control action $a_{t}$ at time step $t$ based on the observation $o_{t}$. We assume that the RL control policy has access to the geometric
  observations related to the closet 10 waypoints in the trajectory. More precisely, let $\bar{o}_{i} = (r_i, \theta_i - \phi_i, \alpha_i-\phi_i, v_i -v)$, be the  geometric
  observations related to waypoint $i$, where $r_i$ is 
  distance between vehicle and the $i$th waypoint, $\theta_i - \phi_i$ is the relative heading of waypoint $i$
  with respect to the vehicle, $\alpha_i - \phi_i$ is the heading error and $v_i - v$ is the velocity
  error of the vehicle with respect to the waypoint $i$. Then $o_{t} = \{\bar{o}_{i(t)+j}\}^{10}_{j=1}$, where $i(t)$ is the next waypoint  at time $t$. Fig. \ref{fig:warobs} shows the geometric observations available at a particular time step.

\begin{figure}
  \vspace*{-3.2cm}
  \resizebox{80mm}{!}{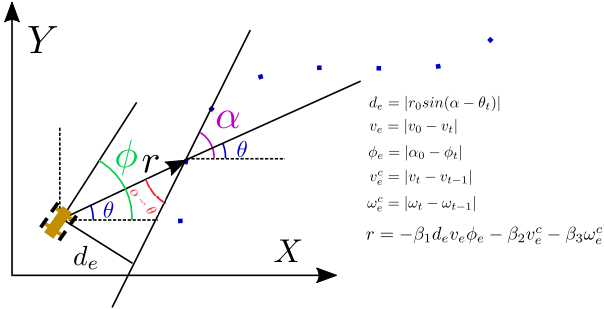}
  \caption{The reward is calculated with respect to the closest waypoint in the trajectory. Here $d_e$ is the corsstrack
  error, $v_e$ is the velocity error, $\phi_e$ is the heading error of the vehicle corresponding to the closest waypoint. $v_e$ and $\omega_e$ are the differences 
  between current and previous linear and angular velocities of the vehicle, respectively.}
  \label{fig:warrew}
\end{figure}

The reward $r_{t}$ received by the RL policy  at time $t$ is computed as
\begin{align*}
     r = -\beta_1d_ev_e\phi_e - \beta_2v_e^c - \beta_3\omega_e^c,
\end{align*}
where $d_e$ is the the crosstrack error, $v_e$ is the velocity error, $\phi_e$ is heading error, and  $v_e^c$ and $\omega_e^c$ are the errors corresponding to the difference between current and previous angular and linear velocities, respectively. All the errors are calculated with respect to the closest waypoint.
Fig. \ref{fig:warrew} gives a geometric interpretation of the reward computation.  The coefficients $\beta_1$, $\beta_2$, $\beta_3$ are selected using hyperparameter tuning.  


\begin{figure}
  \centering{
  \resizebox{75mm}{!}{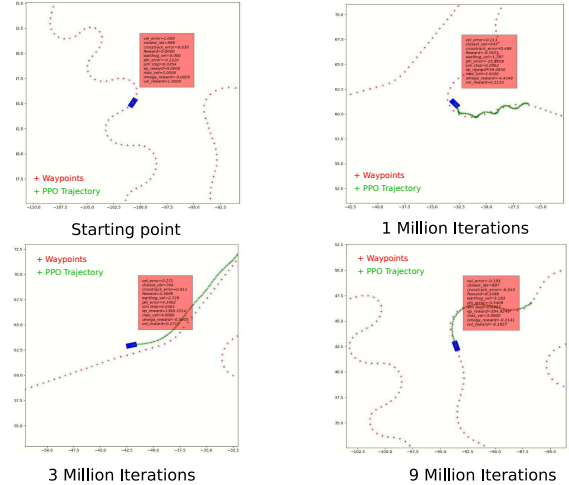}
  \caption{Shows the progression of the baseline RL policy learning in simulation. Red points are the waypoints recorded by driving the
  vehicle in simulation. The blue box shows the current position of the vehicle and green points represent the RL policy trajectory for past 100 time steps. The pink box in the figures display different error terms, rewards, time steps and various simulation parameters.}
  \label{fig:ppo_prog}
  }
\end{figure}

We use the PPO  algorithm  \cite{DBLP:journals/corr/SchulmanWDRK17} to learn the baseline RL policy using the simulation engine described above. We fix the number of steps in an episode and reset the starting position of the vehicle close to a random waypoint in the trajectory at the start of every episode. Fig. \ref{fig:ppo_prog} shows the progress of PPO in simulation for trajectory tracking. It is clear from this 
Fig. that after 9 Million iterations the PPO algorithm  learns good trajectory tracking policy. 


\section{supervised learning-based  sim-to-real transfer}\label{simreal}

\begin{figure}
  \centering{
  \vspace*{-2cm}
  \resizebox{85mm}{!}{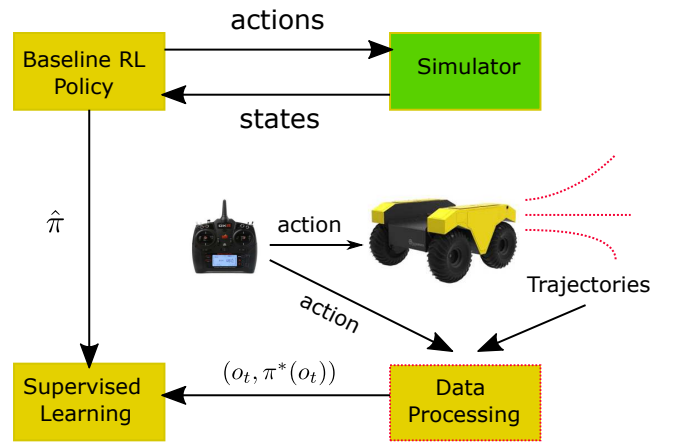}
  \caption{Shows the high level block diagram of our approach. We learn the baseline RL policy in simulation, collect data by driving the vehicle around and use this data to adapt the baseline RL policy to different
  off-road vehicles via supervised learning}
  \label{fig:prop}
  }
\end{figure}

The baseline RL policy cannot be directly transferred to the real-world off-road vehicles because of the minimal and imperfect dynamics model used in the  simulator. Instead of using a naive domain randomization approach to overcome this sim-to-real gap, we propose a novel supervised learning-based approach which adapts the baseline RL policy using the real-world data.  The real-world data is obtained by driving the off-road vehicles (Warthog and Moose) for 30 minutes. The overall approach is represented in Fig. \ref{fig:prop}.



\begin{figure}[H]
  \centering{
  \vspace*{-6cm}
  \resizebox{90mm}{!}{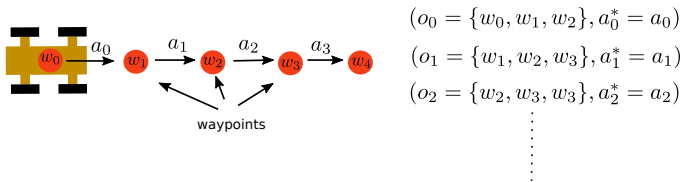}
  \caption{Shows the data collection process on a real vehicle. We discretize the control space and collect waypoints for each combination from this discretization for 5 seconds.}
  \label{fig:datac}
  }
\end{figure}

Consider the real-world data collection from the off-road vehicles, as shown in Fig. \ref{fig:datac}. We start the  vehicle at an arbitrary waypoint $w_0$ and start giving the sequence of actions. The vehicle
executes these actions and moves through waypoints $w_1$, $w_2$, $w_3$ and so on. To collect this data, we discretize the linear velocity range (0 to 5m/s) into 1m/s intervals and angular velocity range (-2 to 2rad/s) into 0.2rad/s intervals and collect 5 seconds of trajectory data for each possible combination of linear and 
angular velocity from this discretization.

\begin{figure}[H]
  \centering{
  \vspace*{-5cm}
  \resizebox{85mm}{!}{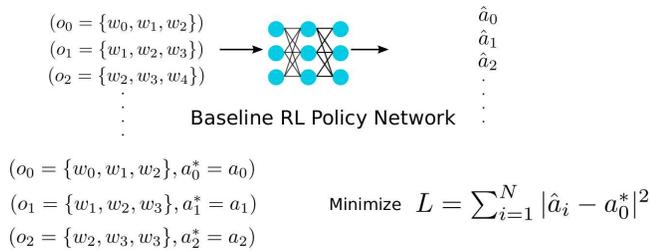}
  \caption{We feed the waypoint data collected using the real vehicle to the  baseline RL policy network and get predicted actions. We then minimize the error 
between the predicted actions and actual actions to update the baseline RL policy network.}
  \label{fig:mean_train}
  }
\end{figure}

To illustrate our supervised learning-based sim-to-real transfer approach which adapts the baseline RL policy to specific vehicle,  assume that the observation consists of the closest 3 waypoints. We  create data pairs of the form $(o_0, a^*_0)$, $(o_1, a^*_1)$, $(o_2, a^*_2)$ and so on from the data collected from the real vehicle. We will also create  data pairs of the form $(o_0, \hat{a}_0)$, $(o_1, \hat{a}_1)$, $(o_2, \hat{a}_2)$ and so on, where $\hat{a}_t = \pi_{\text{RL}}(o_{t})$ is the control action prescribed by the baseline RL policy. Now, the data tuple $(o_{i}, a^*_{i}, \hat{a}_{i})$ can be used as data point for supervised learning, where $o_{i}$ is the input, $a^*_{i}$ is the correct output, $\hat{a}_{i}$ is the estimated output from the baseline RL policy network. We can then update the baseline RL policy network using backpropagation, with the loss function $|a^*_{i} -  \hat{a}_{i}|^{2}$.  We illustrate this procedure in Fig. \ref{fig:mean_train}. 

\section{Results}
We present our results in this section. Please note that in the plots and descriptions, the baseline RL policy (obtained as described in Section 3) is denoted as \textit{vanilla PPO}, to distinguish it from the updated policy obtained after the  supervised learning-based sim-to-real transfer. 

\begin{figure}[H]
   \centering{
   \vspace*{-2.4cm}
   \resizebox{85mm}{!}{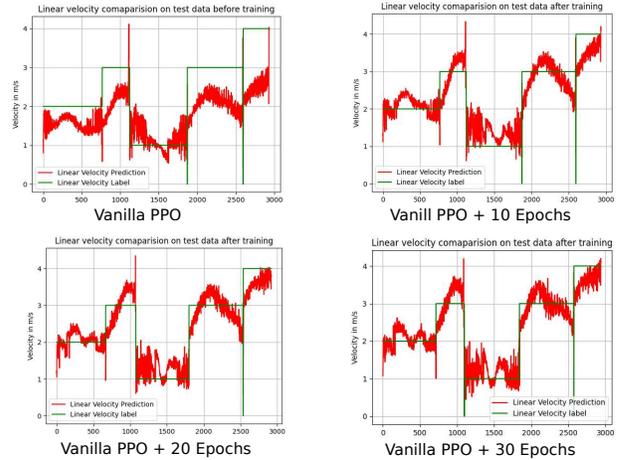}
   \caption{Shows the linear velocity output of vanilla PPO and our method 
   after 10-30 epochs of supervised learning-based sim-to-real transfer. Green curve  is the linear velocity and red curve is the output of the corresponding policy.}
   \label{fig:lin_vel_comp}
   }
 \end{figure}

Fig. \ref{fig:lin_vel_comp}
 plots the optimal linear velocity actions and the linear velocity output of the vanilla PPO algorithm 
 for the data collected during 30 minutes of driving the Warthog. 
 It is clear from this Fig. that the vanilla PPO output 
 is significantly different from the optimal output. After 10 epoch of supervised learning-based sim-to-real transfer,  the linear velocity output improves and we see marginal improvement after 20 and 30 epochs of training.
 
 \begin{figure}[H]
   \centering{
   \vspace*{-1.8cm}
   \resizebox{85mm}{!}{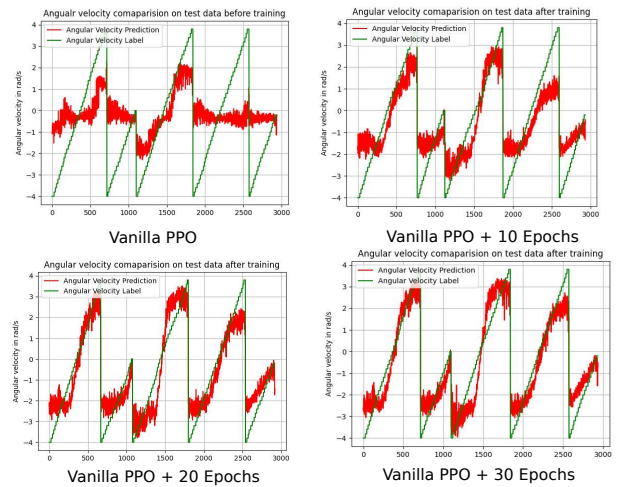}
   \caption{Shows the angular velocity output of vanilla PPO and our method after 10-30 epochs  supervised learning-based sim-to-real transfer. Green curve is the angular velocity label and red curve is the output of the corresponding policy}
   \label{fig:angular_vel_comp}
   }
 \end{figure}
 Fig. \ref{fig:angular_vel_comp}
 plots the optimal angular velocity actions and the angular velocity output of the vanilla PPO algorithm  for the same data collected on Warthog. Again, we see similar results 
 as we observed for linear velocity: vanilla PPO output 
 is significantly different from the optimal output for the angular velocity. After 10 epochs of  supervised learning-based sim-to-real transfer,  the angular velocity 
 output improves. Moreover, it  shows significant improvement after 20 and 30 epochs. 
 
%
%

\begin{figure}[H]
   \centering{
   \vspace*{-2cm}
   \resizebox{70mm}{80mm}{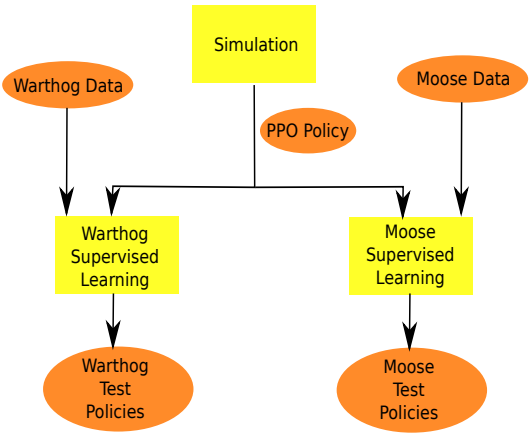}
   \caption{Testing Setup}
   \label{fig:test_setup}
   }
 \end{figure}

 To avoid overfitting and to determine the optimal number of epochs for supervised learning, we test the simulation policy trained for different number of supervised learning epochs,   on both the vehicles in various off-road scenarios and evaluated their performance based on the crosstrack 
 error. Due to the space constraints, we present the results only for two different scenarios for each vehicle. The first scenario has grassy terrain while the second scenario consists of dusty terrain Fig. \ref{fig:terrains}. The testing setup is shown in Fig. \ref{fig:test_setup}. We use PPO simulation policy, train it using the Warthog data for 10, 20, 30 and 40 epochs to get various test policies for Warthog. Similarly, we  train it using the Moose data for 
 the same number of epochs to get test policies for Moose. 
 
  \begin{figure}[H]
   \centering{
   \vspace*{-3cm}
   \resizebox{90mm}{90mm}{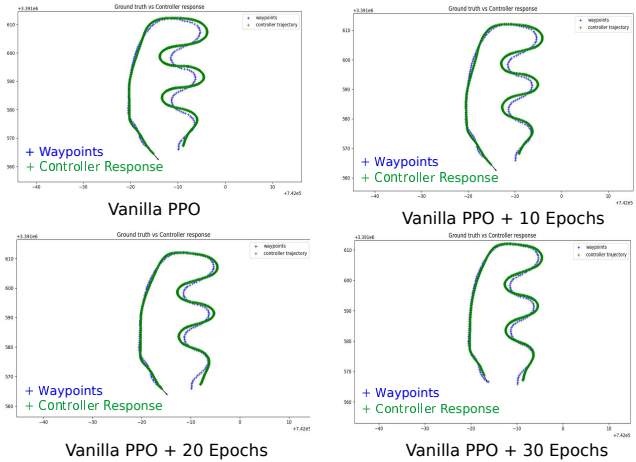}
   \caption{Shows the waypoints for the Warthog test scenario-1 in blue 
   and the trajectory generated by various policies in green.}
   \label{fig:traj_comp}
   }
 \end{figure}
 
\begin{figure}[H]
    \centering{
    \vspace*{-1.7cm}
    \resizebox{90mm}{!}{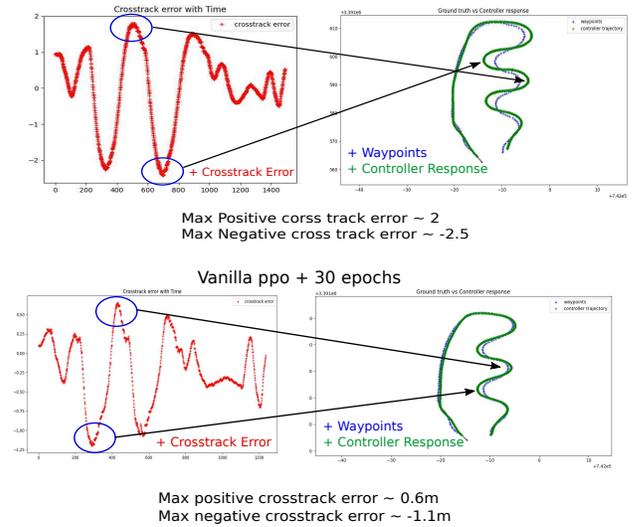}
    \caption{Shows the crosstrack error and trajectory output of two policies on Warthog for test scenario-1. Blue curves show the waypoints and green curves show the trajectory output of the policies. Top row shows the plots for the vanilla PPO and bottom row shows plots for our approach after 30 epochs  supervised learning-based sim-to-real transfer. The blue ellipses show the maximum positive and negative crosstrack errors and the arrow corresponding to each ellipse
    points to the location in the trajectory where the error is observed.}
    \label{fig:cross_comp_traj}
    }
  \end{figure}

\begin{figure}[H]
    \centering{
    \resizebox{70mm}{50mm}{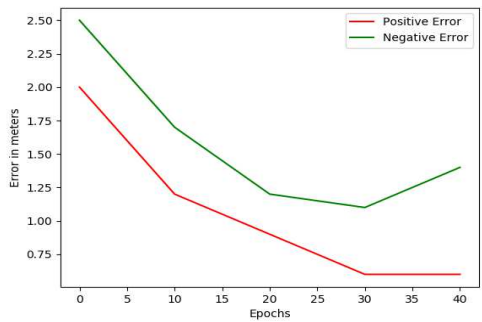}
    \caption{Warthog Crosstrack Error Comparison: Test Run-1}
    \label{fig:cross_comp}
    }
  \end{figure}

 Each scenario consists of a trajectory collected by driving the vehicles at 4m/s. Some of the 
 terrains used for test scenarios is shown in in Fig. \ref{fig:terrains}. Both Warthog and Moose are 
 equipped with a Vectornav-300 GPS which is used to collect the waypoints (50cm spacing) for the trajectories. Fig. \ref{fig:traj_comp} shows the result of running a vanilla PPO algorithm and our proposed supervised learning-based sim-to-real transfer approach  trained for 10, 20 and 30 epochs using  Warthog data for test scenario-1.  
 Fig. \ref{fig:cross_comp_traj} shows the maximum crosstrack errors in positive and negative directions for vanilla PPO
 and our method after 30 epochs of learning  on Warthog data for test scenario-1. Fig. \ref{fig:cross_comp} plots the positive and  negative crosstrack erros with number of epochs. It is clear from these plots that the maximum crosstrack error in  negative direction shows improvement up to 30 epochs of training  and starts increasing after that. We see similar results for  different off-road scenarios for Warthog and hence 30 epochs  is the optimal number of epochs for supervised learning for Warthog. 

\begin{figure}[H]
    \centering{
    \vspace*{-5.2cm}
    \resizebox{90mm}{100mm}{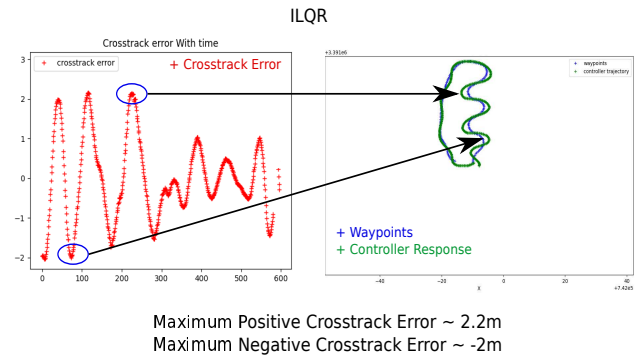}
    \caption{Shows the crosstrack error and trajectory output of ILQR for test scenario-1 on Warthog. Blue curves show the waypoints and green curves show the trajectory output of the ILQR. The blue ellipses show the maximum positive and negative crosstrack errors and the arrow corresponding to each ellipse
    points to the location in the trajectory where the error is observed.}
    \label{ilqr_error}
    }
  \end{figure}

 Fig. \ref{ilqr_error} shows the crosstrack error performance of the ILQR algorithm \cite{DBLP:abs200714492} on test scenario-1 for Warthog. It is clear from this Fig. 
 that the performance of the  ILQR algorithm is  worse than our controller as indicated by the 2.2m maximum absolute crosstrack error.
 
\begin{figure}[H]
    \centering{
    \vspace*{-2cm}
    \resizebox{90mm}{!}{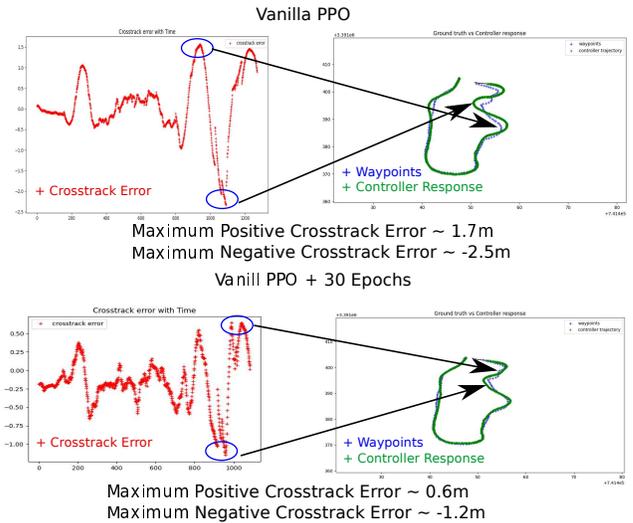}
    \caption{Shows the crosstrack error and trajectory output of two policies on Warthog for test scenarios-2.Blue curves shows the waypoints and green curves show the trajectory output of the policies. Top row shows the plots for the vanilla PPO and bottom row shows the plots for our method after 30 epochs of supervised learning-based sim-to-real transfer.}
    \label{fig:warthog_test2_cross}
    }
  \end{figure}

 Fig. \ref{fig:warthog_test2_cross} shows the trajectroy output and crosstrack errors for test scenario-2, for Warthog, which involves rough terrain and elevations, for vanilla PPO and  our method with 30 epochs of supervised learning. Both of these runs show that our method is able to adapt the vanilla PPO policy trained in simulation to the real Warthog as indicated by the maximum absolute crosstrack error which decreased from
 2.5m to 1.1m for the first scenario and from 2.5m to 1.2m for the second scenario.

\begin{figure}[H]
    \centering{
    \vspace*{-2cm}
    \resizebox{85mm}{!}{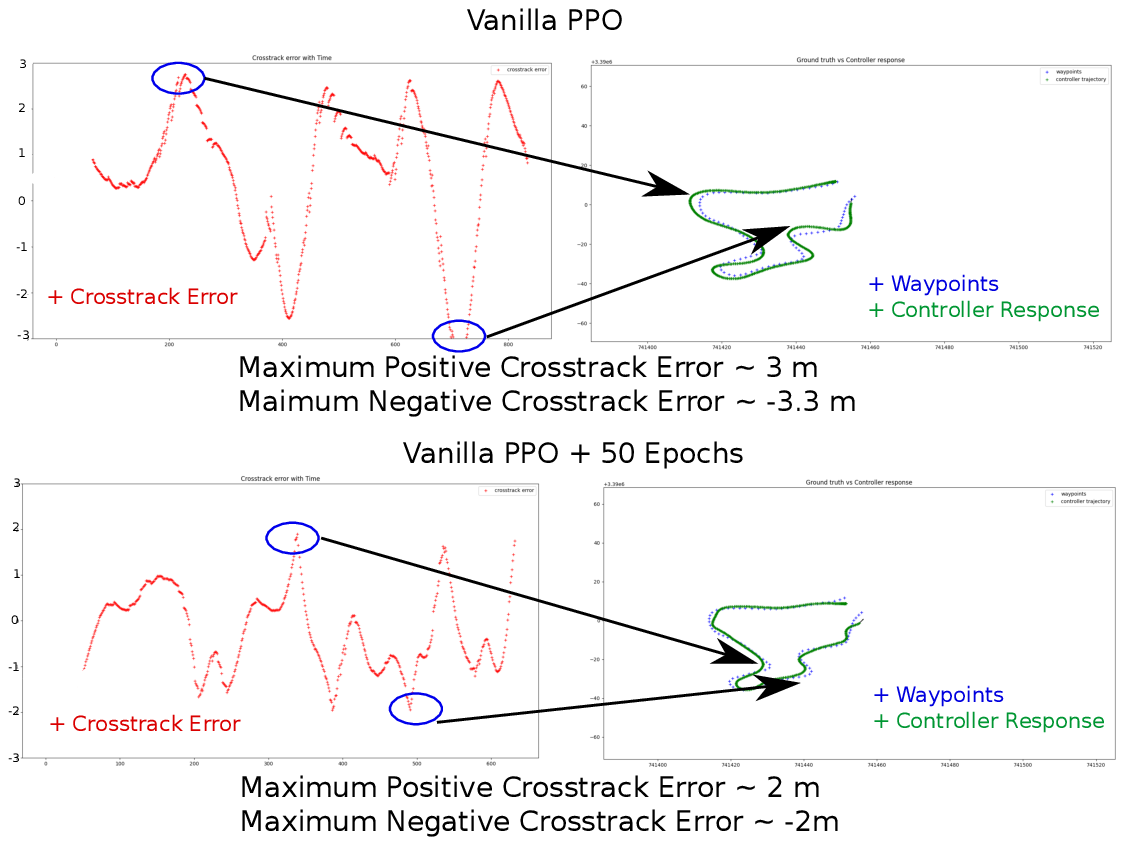}
    \caption{Shows the crosstrack error and trajectory output of two policies on Moose for test scenarios-1. Top row shows the plots for the vanilla PPO and bottom row shows the plots for our method after 50 epochs of supervised learning-based sim-to-real transfer}
    \label{fig:moose_run1cross_comp_traj}
    }
  \end{figure}

\begin{figure}[H]
    \centering{
    \vspace*{-3.4cm}
    \resizebox{90mm}{!}{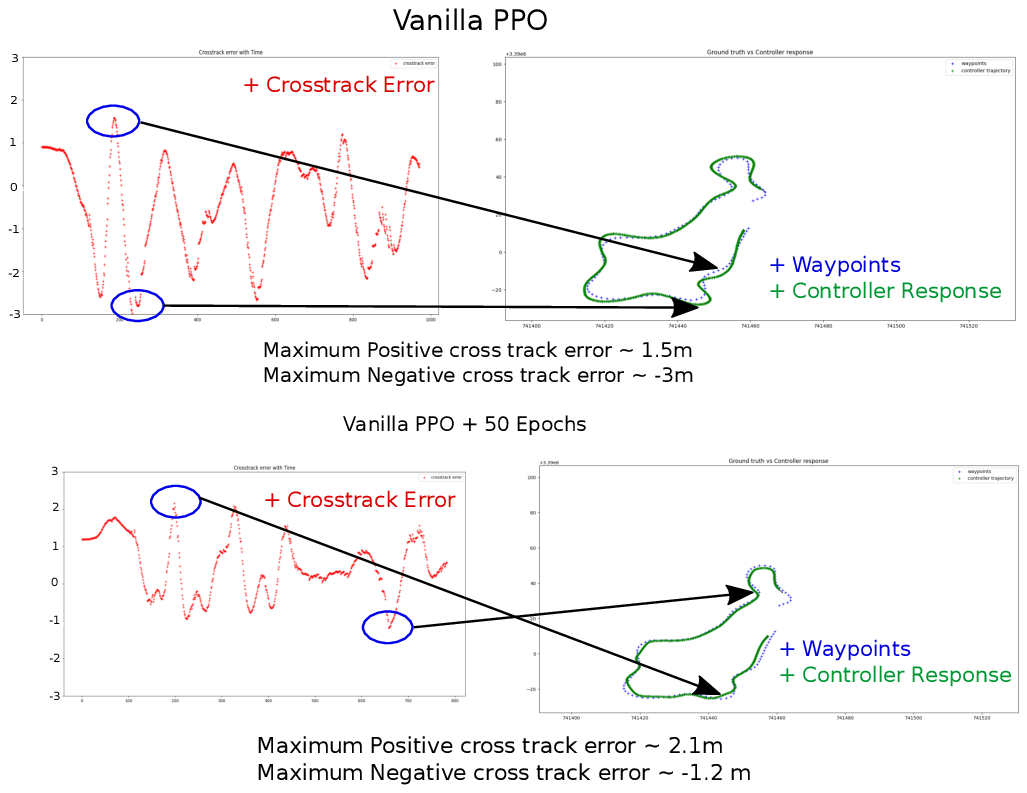}
    \caption{Shows the crosstrack error and trajectory output of two policies on Moose for test scenarios-2. Top row shows the plots for the vanilla PPO and bottom row shows the plots for our method after 50 epochs of supervised learning-based sim-to-real transfer}
    \label{fig:moose_run2cross_comp_traj}
    }
  \end{figure}

 Figures \ref{fig:moose_run1cross_comp_traj} and  \ref{fig:moose_run2cross_comp_traj} plot the performance of the vanilla PPO and our method after 50 epochs of supervised learning-based sim-to-real transfer on Moose for test scenario-1 and test scenairo-2 respectively.  
 The maximum absolute crosstrack errors for the first run decreased from 3m to 2m and from 3m to 2.1m for the test 
 scenarios-1 and 2 respectively. We use the similar method as described above for finding the optimal number of 
 epochs for supervised learning for Moose: after 50 epochs, the maximum absolute crosstrack error starts increasing and hence 50 is the optimal number of epochs for Moose.

 All the above plots  show that our method is able to adapt
 the baseline RL policy learned in simulation  with a very simple model of the vehicles to real-world off-road vehicles with different dynamics.

  \section{Conclusions and Future work}
In this work, we presented a supervised learning-based sim-to-real transfer approach that can adapt a baseline RL policy trained on a simple kinematics simulator to real-world off-road vehicles. We tested and validated our proposed approach in two different off-road vehicles, Warthog and Moose.  We also compared our method with a 
standard model-based controller, namely the ILQR controller for trajectory tracking. We showed that our proposed approach is superior to the existing  approaches in terms of the crosstrack error performance in real-world off-road vehicles. In future, we plan to extend this approach to Ackermann steering off-road vehicles like Polaris Ranger
and use computer vision approaches  to improve the trajectory tracking algorithm further.




  
   
\bibliographystyle{IEEEtran}
\bibliography{IEEEabrv,IEEEexample}

\begin{thebibliography}{10}
\providecommand{\url}[1]{#1}
\csname url@samestyle\endcsname
\providecommand{\newblock}{\relax}
\providecommand{\bibinfo}[2]{#2}
\providecommand{\BIBentrySTDinterwordspacing}{\spaceskip=0pt\relax}
\providecommand{\BIBentryALTinterwordstretchfactor}{4}
\providecommand{\BIBentryALTinterwordspacing}{\spaceskip=\fontdimen2\font plus
\BIBentryALTinterwordstretchfactor\fontdimen3\font minus
  \fontdimen4\font\relax}
\providecommand{\BIBforeignlanguage}[2]{{%
\expandafter\ifx\csname l@#1\endcsname\relax
\typeout{** WARNING: IEEEtran.bst: No hyphenation pattern has been}%
\typeout{** loaded for the language `#1'. Using the pattern for}%
\typeout{** the default language instead.}%
\else
\language=\csname l@#1\endcsname
\fi
#2}}
\providecommand{\BIBdecl}{\relax}
\BIBdecl

\bibitem{mnih2015human}
V.~Mnih, K.~Kavukcuoglu, D.~Silver, A.~A. Rusu, J.~Veness, M.~G. Bellemare,
  A.~Graves, M.~Riedmiller, A.~K. Fidjeland, G.~Ostrovski \emph{et~al.},
  ``Human-level control through deep reinforcement learning,'' \emph{nature},
  vol. 518, no. 7540, pp. 529--533, 2015.

\bibitem{silver2016mastering}
D.~Silver, A.~Huang, C.~J. Maddison, A.~Guez, L.~Sifre, G.~Van Den~Driessche,
  J.~Schrittwieser, I.~Antonoglou, V.~Panneershelvam, M.~Lanctot \emph{et~al.},
  ``Mastering the game of go with deep neural networks and tree search,''
  \emph{nature}, vol. 529, no. 7587, pp. 484--489, 2016.

\bibitem{lillicrap2016continuous}
T.~P. Lillicrap, J.~J. Hunt, A.~Pritzel, N.~Heess, T.~Erez, Y.~Tassa,
  D.~Silver, and D.~Wierstra, ``Continuous control with deep reinforcement
  learning.'' in \emph{ICLR (Poster)}, 2016.

\bibitem{levine2016end}
S.~Levine, C.~Finn, T.~Darrell, and P.~Abbeel, ``End-to-end training of deep
  visuomotor policies,'' \emph{The Journal of Machine Learning Research},
  vol.~17, no.~1, pp. 1334--1373, 2016.

\bibitem{akkaya2019solving}
I.~Akkaya, M.~Andrychowicz, M.~Chociej, M.~Litwin, B.~McGrew, A.~Petron,
  A.~Paino, M.~Plappert, G.~Powell, R.~Ribas \emph{et~al.}, ``Solving rubik's
  cube with a robot hand,'' \emph{arXiv preprint arXiv:1910.07113}, 2019.

\bibitem{haarnoja2019learning}
T.~Haarnoja, S.~Ha, A.~Zhou, J.~Tan, G.~Tucker, and S.~Levine, ``Learning to
  walk via deep reinforcement learning,'' in \emph{Robotics: Science and
  Systems}, 2019.

\bibitem{warthog}
``Warthog, clearpath robotics,''
  \url{https://clearpathrobotics.com/warthog-unmanned-ground-vehicle-robot/}.

\bibitem{moose}
``Moose, clearpath robotics,'' \url{https://clearpathrobotics.com/moose-ugv/}.

\bibitem{sadeghi2017cad}
F.~Sadeghi and S.~Levine, ``{CAD2RL:} real single-image flight without a single
  real image,'' in \emph{Robotics: Science and Systems}, 2017.

\bibitem{tobin2017domain}
J.~Tobin, R.~Fong, A.~Ray, J.~Schneider, W.~Zaremba, and P.~Abbeel, ``Domain
  randomization for transferring deep neural networks from simulation to the
  real world,'' in \emph{2017 IEEE/RSJ international conference on intelligent
  robots and systems (IROS)}.\hskip 1em plus 0.5em minus 0.4em\relax IEEE,
  2017, pp. 23--30.

\bibitem{peng2018sim}
X.~B. Peng, M.~Andrychowicz, W.~Zaremba, and P.~Abbeel, ``Sim-to-real transfer
  of robotic control with dynamics randomization,'' in \emph{2018 IEEE
  international conference on robotics and automation (ICRA)}.\hskip 1em plus
  0.5em minus 0.4em\relax IEEE, 2018, pp. 3803--3810.

\bibitem{80202}
K.~Narendra and K.~Parthasarathy, ``Identification and control of dynamical
  systems using neural networks,'' \emph{IEEE Transactions on Neural Networks},
  vol.~1, no.~1, pp. 4--27, 1990.

\bibitem{nolin2}
S.~Chen, S.~A. Billings, and P.~M. Grant, ``Non-linear system identification
  using neural networks,'' \emph{International Journal of Control}, vol.~51,
  no.~6, pp. 1191--1214, 1990.

\bibitem{IEEE:1383790}
J.~{Kocijan}, R.~{Murray-Smith}, C.~E. {Rasmussen}, and A.~{Girard}, ``Gaussian
  process model based predictive control,'' in \emph{Proceedings of the 2004
  American Control Conference}, vol.~3, June 2004, pp. 2214--2219 vol.3.

\bibitem{IEEEexample:gprl2}
J.~{Ko}, D.~J. {Klein}, D.~{Fox}, and D.~{Haehnel}, ``Gaussian processes and
  reinforcement learning for identification and control of an autonomous
  blimp,'' in \emph{Proceedings 2007 IEEE International Conference on Robotics
  and Automation}, April 2007, pp. 742--747.

\bibitem{IEEEexample:localgp}
D.~{Nguyen-Tuong} and J.~{Peters}, ``Local gaussian process regression for
  real-time model-based robot control,'' in \emph{2008 IEEE/RSJ International
  Conference on Intelligent Robots and Systems}, Sep. 2008, pp. 380--385.

\bibitem{IEEEexample:gpquad}
G.~{Cao}, E.~M. {Lai}, and F.~{Alam}, ``Gaussian process model predictive
  control of unmanned quadrotors,'' in \emph{2016 2nd International Conference
  on Control, Automation and Robotics (ICCAR)}, 2016, pp. 200--206.

\bibitem{IEEEexample:gpdataeff}
M.~Deisenroth, D.~Fox, and C.~Rasmussen, ``Gaussian processes for
  data-efficient learning in robotics and control,'' \emph{IEEE Transactions on
  Pattern Analysis \& Machine Intelligence}, vol.~37, no.~02, pp. 408--423, feb
  2015.

\bibitem{IEEEexample:gprldataeff}
S.~Kamthe and M.~Deisenroth, ``Data-efficient reinforcement learning with
  probabilistic model predictive control,'' in \emph{International conference
  on artificial intelligence and statistics}.\hskip 1em plus 0.5em minus
  0.4em\relax PMLR, 2018, pp. 1701--1710.

\bibitem{DBLP:abs200714492}
\BIBentryALTinterwordspacing
A.~Nagariya and S.~Saripalli, ``An iterative {LQR} controller for off-road and
  on-road vehicles using a neural network dynamics model,'' \emph{CoRR}, vol.
  abs/2007.14492, 2020. [Online]. Available:
  \url{https://arxiv.org/abs/2007.14492}
\BIBentrySTDinterwordspacing

\bibitem{PETERS2008682}
J.~Peters and S.~Schaal, ``Reinforcement learning of motor skills with policy
  gradients,'' \emph{Neural networks}, vol.~21, no.~4, pp. 682--697, 2008.

\bibitem{DBLP:KoberP08}
J.~Kober and J.~Peters, ``Policy search for motor primitives in robotics,''
  \emph{Machine learning}, vol.~84, no. 1-2, pp. 171--203, 2011.

\bibitem{5649089}
P.~Kormushev, S.~Calinon, and D.~G. Caldwell, ``Robot motor skill coordination
  with em-based reinforcement learning,'' in \emph{2010 IEEE/RSJ International
  Conference on Intelligent Robots and Systems}, 2010, pp. 3232--3237.

\bibitem{DBLP:KoberOP11}
J.~Kober, E.~Öztop, and J.~Peters, ``Reinforcement learning to adjust robot
  movements to new situations,'' in \emph{IJCAI}, 2011, pp. 2650--2655.

\bibitem{zakharov2019deceptionnet}
S.~Zakharov, W.~Kehl, and S.~Ilic, ``Deceptionnet: Network-driven domain
  randomization,'' in \emph{Proceedings of the IEEE/CVF International
  Conference on Computer Vision}, 2019, pp. 532--541.

\bibitem{tobin2018domain}
J.~Tobin, L.~Biewald, R.~Duan, M.~Andrychowicz, A.~Handa, V.~Kumar, B.~McGrew,
  A.~Ray, J.~Schneider, P.~Welinder \emph{et~al.}, ``Domain randomization and
  generative models for robotic grasping,'' in \emph{2018 IEEE/RSJ
  International Conference on Intelligent Robots and Systems (IROS)}.\hskip 1em
  plus 0.5em minus 0.4em\relax IEEE, 2018, pp. 3482--3489.

\bibitem{chebotar2019closing}
Y.~Chebotar, A.~Handa, V.~Makoviychuk, M.~Macklin, J.~Issac, N.~Ratliff, and
  D.~Fox, ``Closing the sim-to-real loop: Adapting simulation randomization
  with real world experience,'' in \emph{2019 International Conference on
  Robotics and Automation (ICRA)}.\hskip 1em plus 0.5em minus 0.4em\relax IEEE,
  2019, pp. 8973--8979.

\bibitem{DBLP:journals/corr/BrockmanCPSSTZ16}
\BIBentryALTinterwordspacing
G.~Brockman, V.~Cheung, L.~Pettersson, J.~Schneider, J.~Schulman, J.~Tang, and
  W.~Zaremba, ``Openai gym,'' \emph{CoRR}, vol. abs/1606.01540, 2016. [Online].
  Available: \url{http://arxiv.org/abs/1606.01540}
\BIBentrySTDinterwordspacing

\bibitem{DBLP:journals/corr/SchulmanWDRK17}
\BIBentryALTinterwordspacing
J.~Schulman, F.~Wolski, P.~Dhariwal, A.~Radford, and O.~Klimov, ``Proximal
  policy optimization algorithms,'' \emph{CoRR}, vol. abs/1707.06347, 2017.
  [Online]. Available: \url{http://arxiv.org/abs/1707.06347}
\BIBentrySTDinterwordspacing

\end{thebibliography}

\end{document}